\documentclass[11pt]{article}

\usepackage[final]{acl}

\usepackage[table]{xcolor}
\usepackage{times}
\usepackage{latexsym}
\usepackage{booktabs}
\usepackage{arydshln}
\usepackage[T1]{fontenc}

\usepackage[utf8]{inputenc}

\usepackage{microtype}

\usepackage{inconsolata}

\usepackage{graphicx}
\usepackage{xspace}
\usepackage[most]{tcolorbox}
\usepackage{xcolor}
\usepackage{enumitem}

\newcommand{\ph}[1]{\textcolor{orange!85!black}{\{#1\}}}
\newcommand{\br}[1]{\textcolor{blue!70!black}{[\,#1\,]}}
\newtcolorbox{agentbox}[1]{%
  enhanced,
  breakable,
  colback=white,
  colframe=black,
  boxrule=0.6pt,
  arc=2mm,
  left=6pt,right=6pt,top=6pt,bottom=6pt,
  colbacktitle=black!75,
  coltitle=white,
  fonttitle=\bfseries,
  title=#1
}

\title{CSTrader: A Testbed for Language-Grounded Trading in a Community-Driven Virtual Asset Market}

\author{Yao SHI\thanks{Equal contribution.}\footnotemark[2]\quad
  Kingfung Luo\footnotemark[1]\footnotemark[2]\quad
  Nan Tang\thanks{Hong Kong University of Science and Technology (Guangzhou).}\quad
  Yuyu Luo\footnotemark[2]}

\newcommand{\eat}[1]{}

\usepackage{utfsym} 

\definecolor{green}{RGB}{0,128,0}

\definecolor{yellow}{RGB}{255,200,18}

\newcommand{\bi}{\begin{itemize}}
\newcommand{\ei}{\end{itemize}}

\newcommand{\be}{\begin{enumerate}}
\newcommand{\ee}{\end{enumerate}}
\newcommand{\beqn}{\begin{eqnarray*}}
\newcommand{\eeqn}{\end{eqnarray*}}

\newcommand{\etitle}[1]{\vspace{1mm}\noindent{\underline{\em #1}}}

\begin{document}
\maketitle
\begin{abstract}
Niche asset markets, such as Counter-Strike 2 (CS2) weapon skins, are small, volatile, and heavily driven by community discussions and platform rules. These properties make them hard for traditional quantitative models, but provide an ideal testbed for studying how large language models (LLMs) turn unstructured text into trading actions. We present CSTrader, a multi-agent framework for language-grounded trading in the CS2 skin market. The system first integrates heterogeneous signals from various sources, then uses specialized agents for technical analysis, liquidity, events, and (reversed) sentiment, and finally applies risk control, transaction friction, and portfolio management agents to produce buy, sell, or hold decisions under realistic trading frictions. We build a live-like evaluation environment with real CS2 data from a highly volatile period and evaluate several recent LLM backbones. Across models, CSTrader consistently outperforms both a falling market index (-15.62\%) and simple single-prompt LLM baselines, achieving up to a 7.58\% cumulative return with controlled risk. Ablation studies show that liquidity, reversed sentiment, and transaction friction agents are crucial for turning noisy language signals into stable profits, suggesting that niche, language-driven markets are a useful benchmark for future language-to-action research. Code is available at: \url{https://github.com/IatomicreactorI/CSGOTrading}
\end{abstract}

\section{Introduction}
\label{intro}

\begin{figure*}
    \centering
    \includegraphics[width=.85\linewidth]{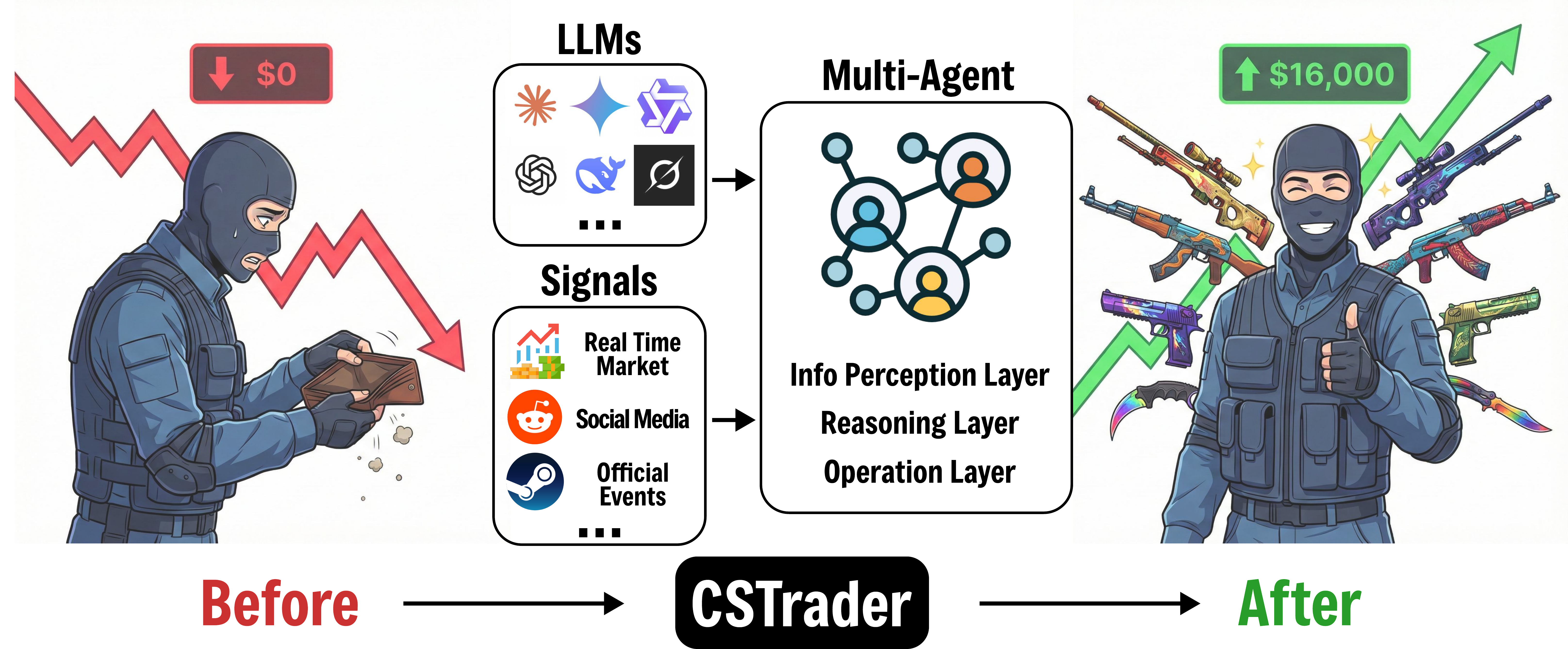}
    \caption{Overview of CSTrader. Our framework takes heterogeneous signals (real-time market prices, social media discussions, and official events) as inputs to a structured multi-agent LLM system, and turns them into trading decisions that significantly improve portfolio performance in the CS2 skin market.}
    \label{fig:teaserconcept}
\end{figure*}

\paragraph{The Appeal of Niche Assets.} Many niche assets provide high investment returns. Examples include Non-Fungible Tokens (NFTs)~\cite{wang2021nft} and collectibles~\cite{valeonti2021collectibles}. These assets usually have high volatility and low trading volumes. Prices depend heavily on news and the behavior of a small group of investors, and they have specific rules such as high transaction fees and ``trade lock'' periods. Skins in the game Counter-Strike 2 (CS2) are a typical example of this asset class~\cite{hardenstein2017skins}, with a large, retail-driven market that remains underanalyzed by institutional investors~\cite{greer2019esports}. An overview of the CS2 skin market, including its scale, asset categories, and trading mechanisms is provided in Appendix~\ref{sec:cs2_market_overview}.

\paragraph{LLMs for Trading.} Recent progress in large language models (LLMs) has changed financial analysis through better document understanding and decision-making capabilities~\cite{liu2023fingpt, wu2024susgen, yu2024fincon}. Traditional machine learning (ML) is good at forecasting time-series data, but LLMs can interpret earnings calls, policy news, and market sentiment. Researchers have used LLMs for financial statements and stock trading~\cite{kim2024financial, li2024alphafin,xiao2024tradingagents}. Multi-agent trading systems have shown good results in several financial markets~\cite{deepfund, berti2025trades, wang2024quantagent, li2025hedgeagents}. However, most of these systems focus on traditional stock markets. This leaves niche assets in real-world trading scenarios mostly ignored. The CS2 market has not received enough attention in LLM-based trading research.

\paragraph{Challenges of Niche Assets Trading.} Trading CS2 skins is different from trading stocks. Most trades are driven by game updates and social-mediated sentiment on platforms like Reddit. These assets often have low liquidity, which means prices can change quickly with small trades~\cite{greer2019esports, hardenstein2017skins, martinelli2017skin}. Effective trading in this market requires a mix of cross-domain reasoning and strategies that are sensitive to specific platform rules. Traditional stock trading approaches often lack these capabilities. At the same time, these properties make the CS2 market an ideal stress test for the ability of models to connect unstructured text, platform rules, and numerical price dynamics in real-time environment. 

\paragraph{Limitations of Existing Methods.} Current approaches have three main limitations when dealing with niche assets.
\begin{itemize}
    \item Traditional ML-based methods focus only on historical price movements. They ignore the ``patch notes'' and community events that drive the CS2 market.
    \item LLM-based methods~\cite{ko2024can, wang2024llmfactor} can analyze events, but they struggle with integrated decision-making. They do not account for the specific rules of niche asset markets, such as high fees or ``trade lock'' periods, which leads to poor behavior in markets like CS2.
    \item Backtesting methods often lack realism. Many systems use historical data in a way that allows ``cheating''. A trading system needs to run in a real-world environment to prove it is effective.
\end{itemize}

\paragraph{Rethinking.} Neither traditional ML nor single LLMs satisfy the three requirements for CS2 trading: processing diverse data, making adaptive decisions under high volatility, and managing portfolio-level risk. These problems are especially serious during periods of market stress. Instead of proposing yet another domain-specific method, we view the CS2 skin market as an idea testbed for language-grounded decision making. Our goal is to build a realistic, online-like evaluation environment in which different LLM-based trading agents can be deployed under the same market frictions and data streams and compared fairly. This perspective leads to the following research question: \textbf{How should we design such an evaluation framework, and what kinds of LLM-based agents can effectively exploit language signals in this setting?}

\paragraph{Our Proposal.} We address this question by building a realistic evaluation framework for language-driven trading, instantiated on the CS2 skin market, and by using it to systematically test different trading agents. An overview of the our CSTrader concept is shown in Figure~\ref{fig:teaserconcept}. Our approach has three components:

\begin{itemize}
    \item \textit{\textbf{Language-driven market as testbed:}} We select the CS2 skin market as a representative language-driven niche asset market, integrate price time series with community and platform text, and expose them as a unified stream of observations that preserves real-world constraints such as fees and trade locks.
    \item \textit{\textbf{Flexible evaluation environment:}} We design an environment interface that is agnostic to the internal architecture of the trading agent. In this work, we instantiate it with single-prompt LLM baselines and a structured multi-agent LLM policy, while keeping the interface extensible to other model classes.
    \item \textit{\textbf{A structured multi-agent LLM policy:}} As a concrete instantiation, we develop CSTrader, a multi-agent LLM policy that decomposes perception, reasoning, and risk-aware portfolio control into interacting agents; CSTrader serves as a strong, interpretable baseline within our framework.
\end{itemize}

Although our framework is instantiated on CS2, the same design applies to other language-driven asset classes such as NFTs and collectibles.

\paragraph{Contributions.} The design of CSTrader and the surrounding framework addresses the unique challenges of language-driven, niche markets and aims to provide insights that go beyond a single application. Our main contributions are:
\begin{itemize}
    \item \textit{\textbf{A realistic evaluation framework for language-grounded trading.}} We frame language-driven markets such as CS2 skins as testbeds for language-grounded decision making and build an environment that streams heterogeneous textual and numerical signals while enforcing real-world frictions (transaction fees and trade-lock rules).
    \item \textit{\textbf{A structured multi-agent LLM baseline.}} Within this framework, we propose CSTrader, a multi-agent LLM trading policy that structures language understanding, market reasoning, and risk control into interacting agents with explicit natural-language rationales.
    \item \textit{\textbf{Superior performance on real-world CS2 market data.}} We conduct experiments on real CS2 skin market data during a highly volatile period and show that our structured multi-agent LLM agents consistently outperform both the overall market index (which declines by \textbf{15.62\%}) and simple single-prompt LLM baselines, achieving up to a \textbf{7.58\%} cumulative return with controlled volatility.
\end{itemize}

\section{Methodology}
\label{method}

\begin{figure*}
    \centering
    \includegraphics[width=.83\linewidth]{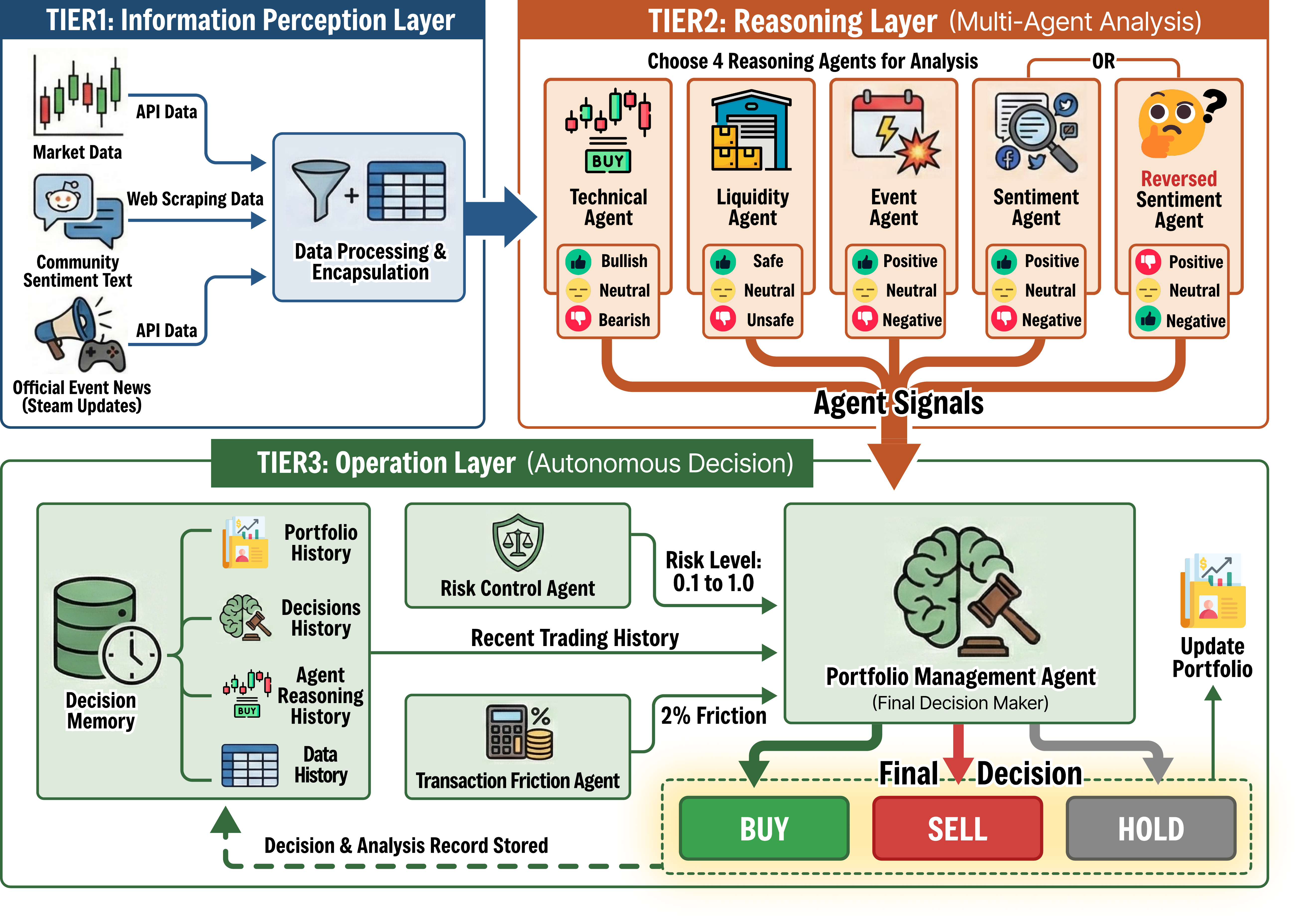}
    \caption{Architecture of CSTrader. The framework is organized into three tiers: TIER1: an Information Perception Layer that aggregates market, community, and official event data into standardized inputs; TIER2: a Reasoning Layer where multiple specialized agents (technical, liquidity, event, sentiment, and reversed sentiment) produce discrete signals with justifications; and TIER3: an Operation Layer that uses risk control, transaction friction, and portfolio management agents with decision memory to integrate these signals into final buy, sell, or hold actions.}
    \label{fig:structure}
\end{figure*}

Our system design takes inspiration from the Theory of Reflexivity introduced by George Soros. This theory suggests that the biases of market participants and the actual market events create a feedback loop where each influences the other. This is particularly relevant for the CS2 skin market because prices are often driven by social trends and community hype rather than just physical utility. By monitoring both the ``market echoes'' on social media and the ``hard data'' from official updates, our framework attempts to navigate this feedback loop.  An overview of the three-tier architecture is shown in Figure~\ref{fig:structure}.

The architecture consists of a three-tier framework. The Information Perception Layer collects and cleans heterogeneous data from multiple sources. The Reasoning Layer employs different agents to analyze the market from technical, social, and event-based perspectives. Finally, the Operation Layer acts as the core of the system by integrating these insights with a memory mechanism to execute trades.

\subsection{Information Perception Layer}
This layer captures diverse data streams to ensure that all trading decisions are based on the most recent market evidence. The collected information is cleaned and transformed into standardized inputs for the Reasoning Layer~\ref{reasoning}.

We categorize the output of this layer into three data types: (1) \textit{market price data}, consisting of daily OHLCV(Open, High, Low, Close, Volume) records for target assets; (2) \textit{community sentiment text}, formed by unstructured discussions from social platforms such as Reddit; and (3) \textit{official event news}, including Steam announcements and game updates that may affect supply and demand. Implementation details and data schemas are provided in the appendix.

\subsection{Reasoning Layer}
\label{reasoning}
This layer uses five specialized agents to provide multi-dimensional analysis. Each agent produces a standardized signal, such as ``Bullish'', ``Bearish'', or ``Neutral'', along with a short natural-language justification.

\etitle{Technical Agent:}
The Technical Agent summarizes price behavior using trend, mean-reversion, momentum, volatility, and volume-level indicators to forecast short- and medium-term movements. Concretely, it combines exponential moving averages, Bollinger-band style mean reversion, RSI-type momentum, volatility regimes, and support/resistance levels; full formulas and hyperparameters are given in the appendix.

\etitle{Sentiment Agent:}
The Sentiment Agent reads Reddit discussions from CS2 trading forums and converts them into a short-term sentiment signal for each skin. It selects high-quality, recent posts for the target ticker and returns a bullish, bearish, or neutral view together with an explanation when data is sparse or noisy.

\etitle{Reversed Sentiment Agent:}
To capture herd effects, the Reversed Sentiment Agent applies a contrarian lens to the standard sentiment output: strong optimism can signal an overheated market, while widespread pessimism can indicate undervaluation. This agent therefore often inverts community mood, helping CSTrader avoid buying into bubbles and exploit overreactions.

\etitle{Liquidity Agent:}
The Liquidity Agent estimates how easily a skin can be traded without large price impact by combining recent trading volume with Reddit engagement. It flags ``liquidity traps''---items with high listed prices but little real demand---and downweights or blocks trades in such assets.

\etitle{Event Agent:}
The Event Agent monitors recent official Steam news and major CS2 updates, focusing on changes to drop pools, item supply, and game visibility. It classifies each news window as bullish, bearish, or neutral for a given skin, indicating whether fundamentals are likely to improve or deteriorate in the near term.

\subsection{Operation Layer}
The Operation Layer translates analytical signals into concrete trades while respecting risk and market frictions.

\etitle{Risk Control Agent:}
The Risk Control Agent aggregates signals from the Reasoning Layer together with the current portfolio state to propose an ``optimal position ratio'' for each skin. It upweights defensive evidence during uncertain periods and downweights aggressive signals, keeping exposure within safe bounds; its full prompt and decision rules are listed in the appendix.

\etitle{Transaction Friction Agent:}
The Transaction Friction Agent encodes platform fees into the decision process. Using the 2\% representative sell fee derived from Appendix Table~\ref{tradingruletb}, it checks whether expected profits remain positive after costs and blocks small, frequent trades that would lose money once fees are applied.

\etitle{Portfolio Management Agent:}
The Portfolio Management Agent is the final decision maker. It combines the suggested position ratio, friction-aware profit check, available cash, and a short decision memory of recent trades to output a discrete action (buy, sell, or hold) and trade size. The agent can override risky recommendations (for example, when prices deviate sharply from recent history) to maintain a stable, consistent strategy over time.

\section{Experiment Setting}
\label{experiment}

\subsection{Setup}
We conduct our experiments using real-world data from the Counter-Strike 2 global official market.

\begin{figure*}
    \centering
    \includegraphics[width=.9\linewidth]{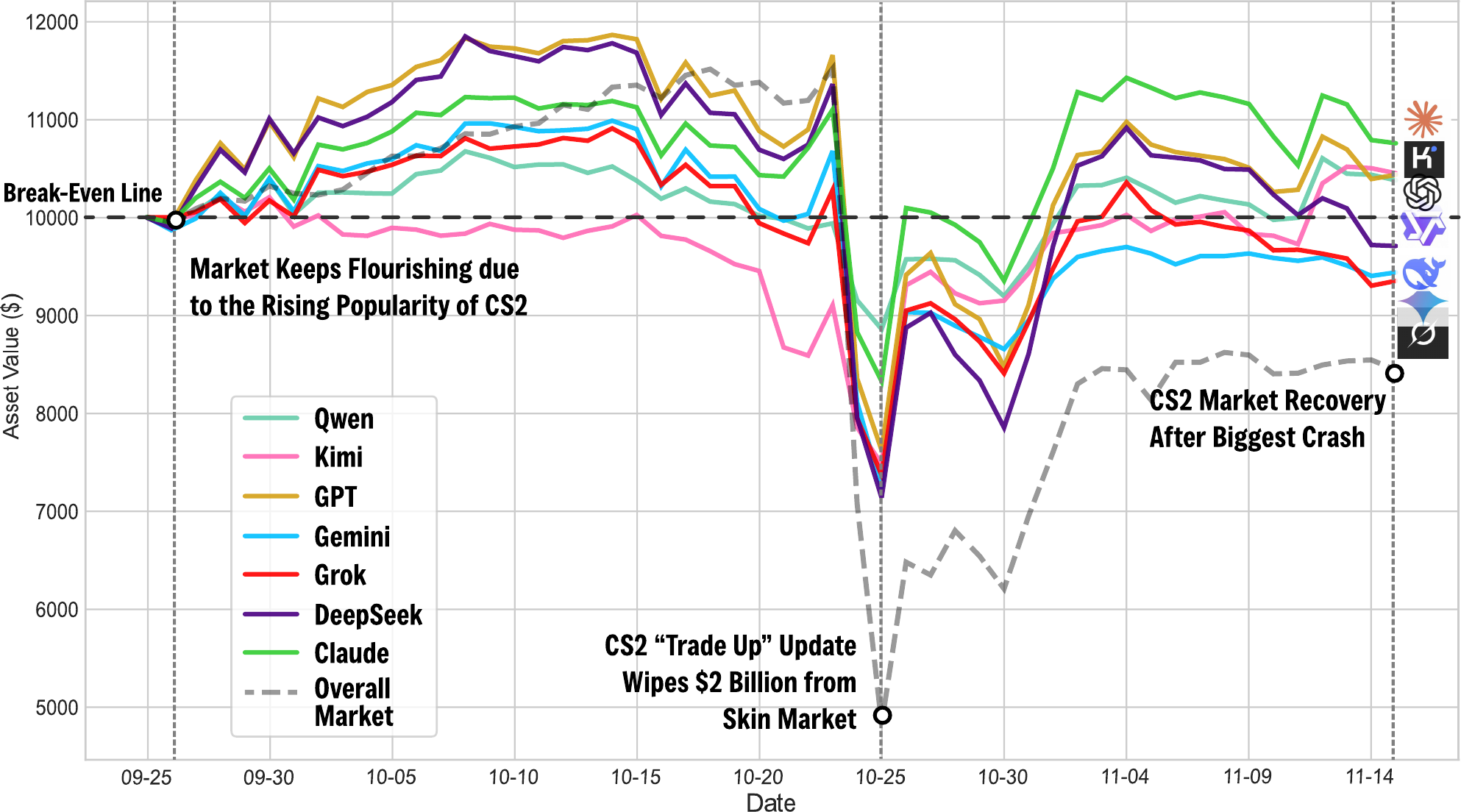}
    \caption{Asset value trajectories of CSTrader with different LLM backbones compared to the overall CS2 market index over the evaluation period. All strategies start from an initial capital of \$10{,}000 (break-even line). The CS2 ``Trade Up'' update on October 25 triggers a sharp market crash that wipes out a large portion of the skin market value, followed by a partial recovery. While the overall market remains below break-even, most CSTrader configurations recover faster and maintain higher asset values.}
    \label{fig:assetvalue}
\end{figure*}

\etitle{Data Acquisition:} 
Our system integrates market data and community events from two main digital sources. We track the whole market through the Steam JSON API to collect daily OHLCV(Open, High, Low, Close, Volume) candles. For community sentiment, the system uses the Reddit API to monitor specific subreddits (such as ``GlobalOffensiveTrade'' and ``csgomarketforum''). Also, since the Reddit API only supports (looking back) from the current time, we use a ``sliding time window'' strategy to align social discussions with specific trading dates.

\etitle{Base Model:} We deploy six of the latest LLMs as our framework's base model from various providers. The knowledge cutoff dates and capabilities of each model are summarized in Table~\ref{modelinfotb}.

\begin{table*}[t!]
  \centering
  \caption{Summary of LLM backbones used as base models in CSTrader. For each provider, we list the model name, release date, knowledge cut-off, Elo rating, and whether the model is open source.}
  \label{modelinfotb}
  \resizebox{.95\textwidth}{!}{
  \begin{tabular}{lccccc}
    \toprule
    \textbf{Providers} & \textbf{Models} & \textbf{Release Date} & \textbf{Knowledge Cut-off} & \textbf{Elo} & \textbf{Open Source}  \\
    \midrule
     DeepSeek & DeepSeek-V3.2~\cite{liu2025deepseek} & Dec 2025 & Aug 2025 & 1421 & ✓ \\
     Alibaba & Qwen-Max~\cite{yang2025qwen3} & Sep 2025 & Jan 2025 & 1441 & ✗ \\
     Kimi & Moonshot-v1~\cite{moonshotv1} & Jan 2025 & Feb 2024 & 1260 & ✗ \\
     Google & Gemini-3-flash-preview~\cite{google2025gemini3flash} & Dec 2025 & Jan 2025 & 1470 & ✗ \\
     OpenAI & GPT-5-mini~\cite{openai2025gpt5} & Aug 2025 & May 2024 & 1443 & ✗ \\
     Anthropic & Claude-sonnet-4~\cite{anthropic2024claude4} & May 2025 & Jan 2025 & 1420 & ✗ \\
     xAI & Grok-4.1-fast-reasoning~\cite{xai2024grok4} & Nov 2025 & Jul 2025 & 1482 & ✗ \\
    \bottomrule
  \end{tabular}
  }
\end{table*}

\etitle{Trading Configuration:} The initial capital for the simulation is \$10,000. The trading period covers a time of high market volatility. During this time, the general CS2 skin market index dropped by 15.62\%. We conduct the experiment during a period of high market volatility, specifically from September 25 to November 15. We use the overall market index~\cite{csqaq_market_index} as our primary baseline to evaluate the effectiveness of the multi-agent system. This index tracks the average price movements of a broad set of assets and serves as a benchmark for passive investment. Its construction is described in Appendix~\ref{sec:csqaq_index}.

\subsection{Evaluation Metrics}
We use several standard metrics~\cite{yu2024fincon, xiong2025flag} to evaluate the performance of each strategy: (1) Cumulative Return (CR)~\cite{hull2012cr}: The total percentage change in the portfolio value; (2) Sharpe Ratio (SR)~\cite{sharpe1994sharpe}: The risk-adjusted return, calculated using a 2\% risk-free rate; (3) Annualized Volatility (AV)~\cite{cochrane1991volatility}: The standard deviation of daily returns, showing price fluctuations; (4) Maximum Drawdown (MDD)~\cite{ang2006downside}: The largest peak-to-trough decline in the portfolio value. (5) Alpha ($\alpha$)~\cite{jensen1968performance}: the excess return that represents the specific profit generated by the model reasoning beyond the overall market. (6) Beta ($\beta$)~\cite{jensen1968performance}: the measure of market risk that compares the volatility of the portfolio to the overall market. We use Alpha ($\alpha$) and Beta ($\beta$) to compare the performance and risk profiles of different model backbones against the overall market. The formal definitions of these metrics are in Appendix~\ref{sec:metrics}.



\begin{table*}[t!]
  \centering
  \caption{Performance of CSTrader with different LLM backbones on the CS2 skin market. We report cumulative return (CR), Sharpe ratio (SR), maximum drawdown (MDD), annualized volatility (AV), beta ($\beta$) relative to the overall market index, and alpha ($\alpha$) capturing excess return over the market.}
  \label{backboneres}
  \begin{tabular}{lcccccc}
    \toprule
    \textbf{Models} & \textbf{CR (\%)} & \textbf{SR} & \textbf{MDD (\%)} & \textbf{AV (\%)} & \textbf{$\beta$} & \textbf{$\alpha$ (\%)}\\
    \midrule
      Claude-sonnet-4 & \cellcolor{green!40} \textbf{7.58} & 0.49 & 25.9 & 79.81 & 0.47 & 66.86 \\
      Moonshot-v1 & \cellcolor{green!30}4.50 & 0.28 & 26.71 & 72.35 & 0.39 & 43.62 \\
      GPT-5-mini & \cellcolor{green!20}4.29 & 0.19 & 35.71 & 99.95 & 0.59 & 54.25 \\
      Qwen-Max & \cellcolor{green!10}3.83 & 0.42 & 17.04 & 38.86 & 0.20 & 28.47 \\
      DeepSeek-V3.2 & \cellcolor{orange!10}-2.89 & -0.16 & 39.74 & 106.10 & 0.65 & 21.71 \\
      Gemini-3-flash-preview &\cellcolor{orange!20} -5.65 & -0.32 & 34.20 & 88.19 & 0.55 & 4.41 \\
      Grok-4.1-fast-reasoning & \cellcolor{orange!30}-6.48 & -0.38 & 32.42 & 82.93 & 0.51 & -1.09 \\
     \hdashline[2pt/2pt]
      Overall Market & \cellcolor{red!40}-15.62 & -0.41 & 58.21 & 145.50 & 1.00 & 0.00 \\
    \bottomrule
  \end{tabular}
\end{table*}

\section{Live Trading Discussion}
\label{data}
We now discuss what the live trading experiments reveal about CS2 as a niche market and about the behavior of different agent and configurations.

\begin{table*}[t!]
  \centering
  \caption{Ablation study on agent combinations for CSTrader with Qwen-Max as the base model. We report cumulative return (CR), Sharpe ratio (SR), maximum drawdown (MDD), and annualized volatility (AV) for different subsets of agents.}
  \label{ablationtb}
  \resizebox{.95\textwidth}{!}{
  \begin{tabular}{lccccc}
    \toprule
    \textbf{Agent Combinations} | \textbf{Base Model: Qwen} & \textbf{CR(\%)} & \textbf{SR} & \textbf{MDD(\%)} & \textbf{AV} \\
    \midrule
    Simple Prompts &\cellcolor{red!40} -11.50 & -0.75 & 26.71 & 65.57 \\
    \hdashline[2pt/2pt]
    Technical + Friction & 1.62 & 0.09 & 19.86 & 50.59 \\
    Technical + Sentiment + Friction & \cellcolor{orange! 40}-10.77 & -1.22 & 18.58 & 37.68 \\
    Technical + Sentiment + Liquidity + Friction  & 0.38 & -0.06 & 11.66 & 29.00 \\
    Technical + Reversed Sentiment + Liquidity + Friction & 5.94 & 0.53 & 19.52 & 54.48 \\
    Technical + Sentiment + Liquidity + Event + Friction & 1.34 & 0.07 & 17.72 & 44.48 \\
    Technical + Reversed Sentiment + Liquidity + No Friction &\cellcolor{yellow!40} 21.96 & 1.66 & 31.06 & 95.04 \\
    Technical + Reversed Sentiment + Liquidity + Event + Friction & 3.83 & 0.42 & 17.04 & 38.86 \\
    \bottomrule
  \end{tabular}
  }
\end{table*}

\begin{figure*}
    \centering
    \includegraphics[width=\linewidth]{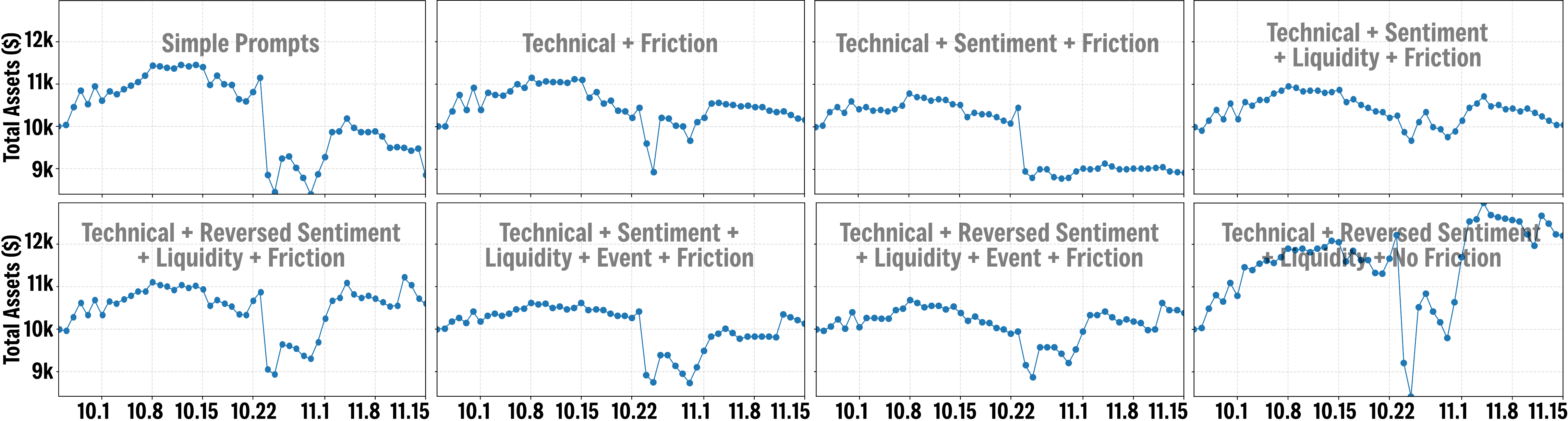}
    \caption{Asset value trajectories of CSTrader under different agent combinations (Qwen-Max backbone). Each panel shows the total portfolio value over time for one set of agents.}
    \label{fig:ablationvalue}
\end{figure*}

\subsection{Performance Against Overall Market}
Our system demonstrates high resilience in a declining market. Figure~\ref{fig:assetvalue} visualizes the portfolio value lines of different LLM backbones and the overall market over the test period. While the CS2 market index fell by 15.62\%, CSTrader configurations based on strong LLMs achieved positive returns and substantially smaller drawdowns, as summarized in Table~\ref{backboneres}. This confirms that the framework can extract ``alpha'' from a niche, event-driven market that is difficult for traditional quantitative models.

The largest shock is the CS2 ``Trade Up'' update on October 25, which causes a sharp market crash followed by a slow, partial recovery of the index. In contrast, most CSTrader portfolios recover faster and stabilize above the break-even line, suggesting that multi-agent reasoning over technical, sentiment, liquidity, and event signals helps navigate sudden regime changes effectively. 

\subsection{Findings for Agents on Niche Market Characteristics}
The CS2 market is a niche category where prices often do not reflect ``actual value'' due to high volatility, thin liquidity, and community bias. The ablation results in Figure~\ref{fig:ablationvalue} and Table~\ref{ablationtb} show that several agents are particularly important.

First, the Liquidity Agent improves the ``realization ability'' of the portfolio by discouraging purchases of items that are easy to list but hard to sell, reflecting the prevalence of ``liquidity traps'' in CS2 skins. Second, the Reversed Sentiment Agent consistently outperforms the Sentiment Agent, indicating a strong herd effect: following Reddit hype tends to buy into bubbles, whereas contrarian signals better capture value opportunities. Third, adding the Event Agent has limited or even negative impact on returns, suggesting that official Steam news is quickly priced in and offers little unique edge. Finally, explicitly modeling a 2\% transaction fee with the Transaction Friction Agent closes the gap between idealized and realistic returns and pushes the system toward less frequent, higher-conviction trades. Together, these observations suggest that success in niche digital asset markets hinges less on reacting to public news and more on counteracting market inefficiencies.

\subsection{Comparison between Base Models}
Finally, we compare different LLM backbones under the same CSTrader framework. Table~\ref{backboneres} shows that stronger reasoning models, such as Claude-sonnet-4, achieve the highest returns and alpha, while more conservative models like Qwen-Max offer the best risk-adjusted stability with low volatility and beta. Aggressive backbones (e.g., DeepSeek-V3.2) can perform well in rising markets but are more exposed to crashes, and weaker models (e.g., Gemini-3-flash-preview) struggle to select profitable items even when timing is approximately correct. Overall, these results suggest that both backbone quality and risk preference shape performance, and that our framework is sensitive enough to reveal these differences.

\section{Related Work}
\label{related}

\subsection{LLMs for Financial Analysis}
Recent studies show that LLMs are becoming useful tools in financial markets. A large group of research projects focuses on investment analysis. For example, FinGPT~\cite{liu2023fingpt} and FinRobot~\cite{yang2024finrobot} transform financial data into research reports. Other models like LLMFactor~\cite{wang2024llmfactor}, XBRL Agent~\cite{han2024xbrl}, and FinSphere~\cite{han2025finsphere} help systems understand specific domains using real-time and multimodal data.

In the field of portfolio management, models like FinMem~\cite{yu2024finmem} and EconAgent~\cite{li2023econagent} use LLMs to make consistent trading decisions. Single-agent frameworks, such as FlagTrader~\cite{xiong2025flag}, QuantAgent~\cite{wang2024quantagent}, and FinAgent~\cite{zhang2024finagent}, focus on creating and executing concrete trading strategies. More recently, multi-agent architectures like TradeAgents~\cite{xiao2024tradingagents}, FinCon~\cite{yu2024fincon}, CryptoTrade~\cite{li-etal-2024-cryptotrade}, and DeepFund~\cite{deepfund} explore how multiple agents can work together to control risks in volatile markets. Although these methods show strong potential for financial analysis and investment, most existing studies remain grounded in traditional stock markets, and niche investment markets are still currently left largely unexplored.

\subsection{Investment Research on Niche Assets}
Beyond traditional equities, a growing body of work studies quantitative trading and risk in niche or alternative asset classes. For cryptocurrencies, recent research links investor attention and multimodal social sentiment to volatility and volume dynamics~\cite{Gadirli2024Investor,liu2025enhancing,roumeliotis2024llms}, and designs multi-agent LLM frameworks for automated crypto portfolio management~\cite{Luo2025LLMPowered}. Other studies extend LLM-based forecasting to carbon credit markets~\cite{Chen2024CanLLM} and bond yields~\cite{walia2025predicting}, or tailor sentiment models to fixed-income news~\cite{Barter2026BondBERT}. Emerging work on illiquid collectibles, such as luxury watches, highlights the diversification and liquidity frictions of these markets~\cite{weisskopf2025time}. Closest to our setting, \citet{10.3389/frai.2025.1702924} build AI-driven algorithmic trading systems for digital game assets in the Counter-Strike~2 skin market, showing that deep models can exploit liquidity and pricing frictions in virtual economies. Together, these studies demonstrate that niche assets, including virtual items, exhibit distinct microstructure and information channels compared with mainstream stocks, motivating specialized, data-driven trading frameworks.

\section{Conclusion}
\label{conclusion}

In this paper, we study CS2 skins as a representative niche, language-driven asset market and use it to build a realistic testbed for language-grounded trading. We design CSTrader, a three-tier multi-agent LLM framework that connects heterogeneous signals from market data, community discussions, and official events to concrete buy, sell, and hold decisions under real-world trading frictions.

Our experiments on real CS2 market data show that CSTrader is robust in a volatile and declining market. Across several strong LLM backbones, our system consistently outperforms the overall market index and simple single-prompt LLM baselines, achieving positive cumulative returns and high alpha while controlling risk. Ablation studies further reveal that agents for liquidity, reversed sentiment, and transaction friction are crucial for turning noisy language signals into stable profits.

This work suggests that niche, language-driven markets are a useful testbed for studying how LLMs turn text into long-horizon financial actions. In future work, we plan to extend our framework to other virtual asset markets, integrate classical quantitative models as additional agents, and explore live trading deployments to better understand how language-based decision systems behave under continuously changing conditions.

\section{Limitation and Future Work}

Despite these promising results, our study has several limitations. First, we focus on a single virtual asset market (CS2 skins) and a liquidity-filtered subset of items, so the observed behaviors may not directly transfer to larger or regulated financial markets. Second, our evaluation is based on historical backtests over a relatively short, highly volatile window around the ``Trade Up'' update, without explicitly modeling order-book dynamics, slippage, or execution latency beyond a simple 2\% transaction fee. Third, CSTrader currently relies on prompt-engineered agents built on top of a small set of frontier LLM backbones; we do not fine-tune models or systematically search over agent architectures and hyperparameters. Finally, while we compare against naive single-prompt LLMs and a passive market index, we do not include strong classical quantitative or reinforcement-learning trading baselines.

These limitations open up several directions for future work. One direction is to extend CSTrader to other virtual economies and small-cap financial instruments, and to run longer-horizon evaluations that cover multiple market regimes. Another is to enrich the environment with more realistic microstructure, including order-book snapshots, execution constraints, and alternative fee schedules, to better approximate live trading conditions. On the modeling side, we plan to explore learning-based agents that combine LLM reasoning with trainable decision layers or offline reinforcement learning, as well as hybrid systems that integrate classical quantitative signals as first-class agents. Finally, we would like to study human-in-the-loop and interpretability aspects of multi-agent LLM traders, for example by analyzing failure cases, agent interactions, and explanation quality when strategies are deployed in or near live markets.


\bibliography{reference}

\appendix
\newpage

\section{CS2 Skin Market Overview}
\label{sec:cs2_market_overview}

The CS2 skin market is one of the largest and most mature virtual-asset economies worldwide and is often described by players as an ``electronic stock market''. Skins are purely cosmetic items that change the appearance (and sometimes audio) of weapons, knives, gloves, or agents without affecting gameplay statistics such as damage or accuracy, which preserves competitive fairness. Each individual skin instance is characterized by continuous attributes such as float value and pattern index, so that even two copies of the ``same'' skin can differ visually and in market price.

Community trackers (e.g., PriceEmpire) estimate that, by late 2025, the notional market capitalization of tradable CS2 skins is on the order of 4--5 billion USD. Valve is reported to earn more than one billion USD per year from key sales and marketplace fees. The ecosystem contains several hundred thousand distinct items and a daily active player base between roughly 600K and 1.5M, with peak concurrent players exceeding 1.8M. The market is dominated by retail participants---ordinary players and collectors---with limited institutional participation, leading to high volatility, sentiment-driven flows, and pronounced liquidity frictions.

\subsection{Item Attributes and Categories}

CS2 skins are organized along several key axes that jointly determine their value:

	\textbf{Rarity.} Item rarity is discretized into tiers that roughly correspond to color codes in the user interface, from low to high: Consumer Grade (white), Industrial Grade (light blue), Mil-Spec (blue), Restricted (purple), Classified (pink), Covert (red), and Extremely Rare (gold, mostly knives and gloves). Higher rarity implies lower drop probability and typically higher price.

	\textbf{Wear level and float value.} Visual wear is divided into five canonical categories: Factory New (FN), Minimal Wear (MW), Field-Tested (FT), Well-Worn (WW), and Battle-Scarred (BS). Underneath, each skin instance has a continuous float value in the range $[0.00, 1.00]$, with lower values corresponding to less visible wear. Market prices are highly sensitive to float, especially for FN and MW items near the lower end of the range.

	\textbf{Pattern index.} For many skins (e.g., Case Hardened, Fade, Crimson Web), the game samples a random pattern index that determines the exact texture overlay. Certain pattern configurations (such as all-blue ``Blue Gem'' Case Hardened knives) are extremely rare and command large premiums over otherwise similar items.

	\textbf{Asset categories.} The core tradable assets are weapon skins for popular rifles, pistols, and sniper rifles such as AK-47, M4A1-S, and AWP, which form the backbone of many portfolios. Gold-tier knives and gloves, with an overall drop probability of roughly 0.26\%, act as high-value pillars of the market, with individual items ranging from hundreds to tens of thousands of USD. Stickers and capsules---especially early tournament series such as Katowice 2014---are highly speculative collectibles, with some individual stickers valued at tens or even hundreds of thousands of USD. Additional categories include agent skins (which change the player model), sprays, weapon cases, keys, medals, and newer cosmetic features such as charms and temporary rental-style access.

\subsection{Acquisition Mechanics}

Players obtain skins via several mechanisms that have direct implications for supply dynamics and player behavior:

	\textbf{Weekly drops.} Eligible accounts with Prime Status receive a ``weekly drop'' when they level up once per week. The drop presents a small menu of items, typically including a weapon case, a low-tier skin, and graffiti options, from which the player can select two items. This mechanism provides a slow, steady inflow of low- to mid-tier items.

	\textbf{Case opening (gacha).} The dominant mechanism for injecting new high-value items is opening weapon cases. Players must purchase keys (priced around 2.5~USD) to open cases obtained via drops or the market. Valve publicly states approximate rarity probabilities: Covert (red) items have a drop chance on the order of 0.64\%, while gold-tier knives and gloves collectively have a probability around 0.26\%. From a financial perspective, case opening is a negative-expected-value gamble, but the possibility of extremely rare, high-value drops sustains very large aggregate volume.

	\textbf{Trade-up contracts.} Players can exchange ten skins of the same rarity tier for one skin of the next-higher tier via an in-game ``trade-up contract''. In a 2025 update, Valve introduced a new form of trade-up that allows combining a set of Covert (red) skins to obtain knives or gloves. This change temporarily drove up demand and prices for certain Covert items and caused sharp dislocations in the knife and glove markets, illustrating how rule changes propagate through the ecosystem.

\subsection{Market Size, Venues, and Frictions}

The primary trading venue is the official Steam Community Market. Trades are denominated in regional fiat currencies but settled into non-withdrawable Steam Wallet balances, meaning that realized profits can only be reinvested within the Steam ecosystem. Steam charges a fee of roughly 15\% on each sale (around 10\% to Valve and 5\% to the game developer), which makes high-frequency trading expensive compared with traditional brokerage accounts.

To combat fraud, money laundering, and automated arbitrage, Valve imposes strict trade-lock rules. Items obtained via trades, market purchases, or case openings cannot be traded again for seven days. In practice, certain sequences of transactions yield an effective holding period close to T+14. Accounts without Steam Guard mobile authentication face even longer delays (up to 15 days) before trades are confirmed. These frictions slow portfolio turnover and create a gap between mark-to-market valuations and realizable proceeds.

Outside the official marketplace, a network of third-party peer-to-peer (P2P) platforms provides alternative liquidity and cash-out channels. In mainland China, NetEase BUFF is the dominant marketplace, enabling RMB cash withdrawals and charging relatively low fees (typically 1--3\%). Internationally, platforms such as Skinport, CS.MONEY, CSFloat, and Tradeit.gg match buyers and sellers via Steam trade offers, sometimes with support for credit cards or cryptocurrencies. Very high-end items (e.g., rare souvenir AWP Dragon Lore or unique pattern knives) are often traded privately through brokers or OTC-style deals rather than public listings.

In 2025 Valve also introduced a rental-like cosmetic access system, in which players can temporarily use certain skins by paying a small fixed fee (roughly the price of a key). This mechanism lowers the barrier to experiencing premium cosmetics while preserving scarcity of permanent ownership, further enriching the ecosystem of use cases around skins.

\subsection{Risks and Market Dynamics}

The CS2 skin market is exposed to several sources of risk that directly influence trading strategies:

	\textbf{Update and policy risk.} Game-balance patches, new case releases, and rule changes (such as modifications to trade-up contracts or trade-lock policies) can cause sudden repricing of whole item classes. For example, adjustments to trade-up rules have previously led to multi-hundred-million-USD swings in estimated market capitalization within days.

	\textbf{Price volatility and liquidity.} Many items are thinly traded, with large gaps between listing prices and actual executed trades. Liquidity is highly skewed toward a small subset of meta-relevant skins and iconic collectibles, while long tails of items exhibit ``liquidity traps'' where nominal prices are high but buy-side depth is shallow.

	\textbf{Security and regulatory risk.} Common threats include phishing websites, API token hijacking, and impersonation scams that exploit Steam's trading workflows. In addition, national regulations on loot boxes and virtual-item gambling (e.g., restrictions in certain European countries) can affect case-opening volumes and platform policies, with second-order effects on prices.

Overall, the CS2 skin ecosystem functions as a large-scale digital collectibles market with a rich set of microstructural frictions, heterogeneous venues, and strongly language-driven information flows. These properties motivate our choice of CS2 as a challenging testbed for language-grounded trading agents.

\section{Agent Instruction}
\subsection{Prompt Template}

\begin{agentbox}{Planner Agent}
You are a planner agent that decides which analysts to perform based on the your knowledge of the ticker and features of analysts.

\medskip
Here is the ticker:
\begin{itemize}[label=-, leftmargin=1.2em, itemsep=2pt, topsep=2pt]
  \item \ph{ticker}
\end{itemize}

\medskip
Here are the available analysts:
\begin{itemize}[label=-, leftmargin=1.2em, itemsep=2pt, topsep=2pt]
  \item \ph{analysts}
\end{itemize}

\medskip
You must provide your decision as a structured output with the following fields:
\begin{itemize}[label=-, leftmargin=1.2em, itemsep=2pt, topsep=2pt]
  \item analysts: selected analyst\_name list
  \item justification: brief explanation of your selection
\end{itemize}
\end{agentbox}

\begin{agentbox}{Technical Agent}
You are a technical analyst evaluating items in CS2 market using multiple technical analysis strategies.

\medskip
The following signals have been generated from our analysis:

\medskip
\textbf{Price Trend Analysis}:
\begin{itemize}[label=-, leftmargin=1.2em, itemsep=2pt, topsep=2pt]
  \item \textit{Trend Following}: \ph{analysis[trend]}
\end{itemize}

\medskip
\textbf{Mean Reversion and Momentum}:
\begin{itemize}[label=-, leftmargin=1.2em, itemsep=2pt, topsep=2pt]
  \item \textit{Mean Reversion}: \\\ph{analysis[mean\_reversion]}
  \item \textit{RSI}: \ph{analysis[rsi]}
  \item \textit{Volatility}: \ph{analysis[volatility]}
\end{itemize}

\medskip
\textbf{Volume Analysis}: \ph{analysis[volume]}

\medskip
\textbf{Support and Resistance Levels}: \\\ph{analysis[price\_levels]}

\medskip
Provide structured output:
\begin{itemize}[label=-, leftmargin=1.2em, itemsep=2pt, topsep=2pt]
  \item signal: \\\br{\texttt{"Bullish"}, \texttt{"Bearish"}, \texttt{"Neutral"}}
  \item justification: Brief explanation of your analysis
\end{itemize}
\end{agentbox}

\begin{agentbox}{Sentiment Analyst}
You are a sentiment analyst evaluating items in CS2 market based on Reddit discussions.

\medskip
Analyze Reddit discussions for \ph{ticker} (\ph{post\_count} posts).
\begin{itemize}[label=-, leftmargin=1.2em, itemsep=2pt, topsep=2pt]
  \item \textit{Direct posts:} price trends, demand/supply factors
  \item \textit{General posts:} overall market mood $\rightarrow$ infer impact on \ph{ticker}
  \item Focus on content sentiment, not just upvotes/comments
  \item If posts $< 5$: return \texttt{"Neutral"} and explain data limits
\end{itemize}

\medskip
Reddit discussions:
\begin{itemize}[label=-, leftmargin=1.2em, itemsep=2pt, topsep=2pt]
  \item \ph{reddit\_posts}
\end{itemize}

\medskip
Give a short-term (1-2 weeks) sentiment: \texttt{Bullish} / \texttt{Bearish} / \texttt{Neutral}.

\medskip
Provide structured output:
\begin{itemize}[label=-, leftmargin=1.2em, itemsep=2pt, topsep=2pt]
  \item signal: \\\br{\texttt{"Positive"}, \texttt{"Negative"}, \texttt{"Neutral"}}
  \item justification: Brief explanation of your analysis
\end{itemize}
\end{agentbox}

\begin{agentbox}{Reversed Sentiment Agent}
You are a contrarian sentiment analyst for CS2 market items. Apply reverse sentiment analysis based on the contrarian hypothesis.

\medskip
Original sentiment signal: \ph{original\_signal}

Original justification: \ph{original\_justification}

\medskip
\textbf{Contrarian Hypothesis:}
\begin{itemize}[label=-, leftmargin=1.2em, itemsep=2pt, topsep=2pt]
  \item Overly bullish Reddit chatter can signal market overheating $\rightarrow$ potentially bearish
  \item Negative chatter can indicate overselling $\rightarrow$ potentially bullish
  \item Neutral sentiment remains neutral
\end{itemize}

\medskip
\textbf{Your task:}
\begin{itemize}[label=-, leftmargin=1.2em, itemsep=2pt, topsep=2pt]
  \item Reverse the signal direction (Bullish $\rightarrow$ Bearish, Bearish $\rightarrow$ Bullish, Neutral $\rightarrow$ Neutral)
  \item Provide a justification explaining the contrarian interpretation
\end{itemize}

\medskip
Evaluate the reversed sentiment for \ph{ticker} based on the contrarian hypothesis.

\medskip
Provide structured output:
\begin{itemize}[label=-, leftmargin=1.2em, itemsep=2pt, topsep=2pt]
  \item signal: \\\br{\texttt{"Positive"}, \texttt{"Negative"}, \texttt{"Neutral"}}
  \item justification: Brief explanation of your analysis
\end{itemize}
\end{agentbox}

\begin{agentbox}{Event Agent}
You are an event analyst for CS2 items. Analyze Steam news for price impact on \{ticker\}.

\medskip
\textbf{Impact Assessment (priority order):}
\begin{enumerate}[leftmargin=1.4em, itemsep=2pt, topsep=2pt]
  \item \textit{Supply mechanism (strongest):} Drop pool, crate/box, rarity, trade-up path changes
  \item \textit{Visibility/popularity (moderate):} New crates, team stickers, weapon balance changes
  \item \textit{Market sentiment (indirect):} Player influx, major updates, speculative activity
\end{enumerate}

\medskip
\textbf{Signal:}
\begin{itemize}[label=-, leftmargin=1.2em, itemsep=2pt, topsep=2pt]
  \item \textit{Bullish:} Increases scarcity/visibility or positive sentiment
  \item \textit{Bearish:} Increases supply, decreases visibility, or negative sentiment
  \item \textit{Neutral:} No clear impact, insufficient data (\ph{news\_count} items), or mixed signals
\end{itemize}

\medskip
\textbf{Steam News (\ph{news\_count} items):}
\begin{itemize}[label=-, leftmargin=1.2em, itemsep=2pt, topsep=2pt]
  \item \ph{steam\_news}
\end{itemize}

\medskip
Evaluate event impact for short-term (1--2 weeks) price movement of \ph{ticker}.
Specify which news items and factors influenced your signal.

\medskip
Provide structured output:
\begin{itemize}[label=-, leftmargin=1.2em, itemsep=2pt, topsep=2pt]
  \item signal: \\\br{\texttt{"Positive"}, \texttt{"Negative"}, \texttt{"Neutral"}}
  \item justification: Brief explanation of your analysis
\end{itemize}
\end{agentbox}

\begin{agentbox}{Liquidity Analyst (CS2 Market)}
You are a liquidity analyst for CS2 items. Analyze liquidity based on trading volume and Reddit engagement.

\medskip
\textbf{Analysis:}
\begin{itemize}[label=-, leftmargin=1.2em, itemsep=2pt, topsep=2pt]
  \item \ph{trading\_volume\_analysis}
  \item \ph{reddit\_engagement\_analysis}
\end{itemize}

\medskip
\textbf{Thresholds:}
\begin{itemize}[label=-, leftmargin=1.2em, itemsep=2pt, topsep=2pt]
  \item \textbf{Volume:} High $\geq$ \ph{volume\_high}, Low $<$ \ph{volume\_low}
  \item \textbf{Reddit:} High (score $\geq$ \ph{reddit\_high\_score} \textbf{or} comments $\geq$ \ph{reddit\_high\_comments}), \\
        Low (score $<$ \ph{reddit\_low\_score} \textbf{and} comments $<$ \ph{reddit\_low\_comments})
  \item \textbf{Min posts:} \ph{reddit\_min\_posts}
\end{itemize}

\medskip
\textbf{Signal:}
\begin{itemize}[label=-, leftmargin=1.2em, itemsep=2pt, topsep=2pt]
  \item \textbf{Bullish:} High volume \textbf{OR} strong engagement (both $\rightarrow$ higher confidence)
  \item \textbf{Bearish:} Low volume \textbf{OR} weak engagement (both $\rightarrow$ higher confidence)
  \item \textbf{Neutral:} Mixed/conflicting indicators or insufficient data
\end{itemize}

\medskip
Evaluate liquidity (bullish/bearish/neutral) for \ph{ticker}. Explain which indicators contributed most.

\medskip
Provide structured output:
\begin{itemize}[label=-, leftmargin=1.2em, itemsep=2pt, topsep=2pt]
  \item signal: \br{\texttt{"Safe"}, \texttt{"Unsafe"}, \texttt{"Neutral"}}
  \item justification: Brief explanation of your analysis
\end{itemize}
\end{agentbox}

\begin{agentbox}{Portfolio Management Agent}
You are a portfolio manager making final trading decisions based on decision memory and the provided optimal position ratio.

\medskip
Decision memory:
\begin{itemize}[label=-, leftmargin=1.2em, itemsep=2pt, topsep=2pt]
  \item \ph{decision\_memory}
\end{itemize}

\medskip
Current Price: \ph{current\_price}\\
Holding Shares: \ph{current\_shares}\\
Tradable Shares: \ph{tradable\_shares}

\medskip
\textbf{Rules:}
\begin{itemize}[label=-, leftmargin=1.2em, itemsep=2pt, topsep=2pt]
  \item If \texttt{tradable\_shares} $>$ 0: you may buy.
  \item If \texttt{tradable\_shares} $<$ 0: you may sell.
  \item If \texttt{tradable\_shares} $\approx$ 0: choose \texttt{Hold}.
\end{itemize}

\medskip
You must provide your decision as a structured output with the following fields:
\begin{itemize}[label=-, leftmargin=1.2em, itemsep=2pt, topsep=2pt]
  \item action: One of \br{\texttt{"Buy"}, \texttt{"Sell"}, \texttt{"Hold"}}
  \item shares: Number of shares to buy or sell, set 0 for hold
  \item price: The current price of the ticker
  \item justification: Briefly explain your decision
\end{itemize}

\medskip
Your response should be well-reasoned and consider all aspects of the analysis.
\end{agentbox}

\begin{agentbox}{Transaction Friction Agent}
You are a portfolio manager making final trading decisions based on decision memory and the provided optimal position ratio.

\medskip
Decision memory:
\begin{itemize}[label=-, leftmargin=1.2em, itemsep=2pt, topsep=2pt]
  \item \ph{decision\_memory}
\end{itemize}

\medskip
Current Price: \ph{current\_price}\\
Holding Shares: \ph{current\_shares}\\
Tradable Shares: \ph{tradable\_shares}

\medskip
Trading friction: selling fee \ph{transaction\_fee\_rate\_pct:.2f}\% (applies to sells only).

\medskip
\textbf{Rules:}
\begin{itemize}[label=-, leftmargin=1.2em, itemsep=2pt, topsep=2pt]
  \item If \texttt{tradable\_shares} $>$ 0: you may buy (no fee on buy).
  \item If \texttt{tradable\_shares} $<$ 0: you may sell; ensure expected downside risk outweighs sell fee.
  \item If \texttt{tradable\_shares} $\approx$ 0 or expected gain $<$ sell-fee impact: choose \texttt{Hold}.
  \item Ensure expected profit after (sell) fees is positive; otherwise \texttt{Hold}.
\end{itemize}

\medskip
You must provide your decision as a structured output with the following fields:
\begin{itemize}[label=-, leftmargin=1.2em, itemsep=2pt, topsep=2pt]
  \item action: One of \br{\texttt{"Buy"}, \texttt{"Sell"}, \texttt{"Hold"}}
  \item shares: Number of shares to buy or sell, set 0 for hold
  \item price: The current price of the ticker
  \item justification: Briefly explain your decision, explicitly noting how the 2\% sell fee impacted the choice
\end{itemize}

\medskip
Your response should be well-reasoned and consider all aspects of the analysis.
\end{agentbox}

\begin{agentbox}{Risk Control Agent}
You are a professional risk control analyst.
Please evaluate the risk of the ticker and set the optimal position ratio based on analyst signals and portfolio state.

\medskip
Here are the analyst signals:
\begin{itemize}[label=-, leftmargin=1.2em, itemsep=2pt, topsep=2pt]
  \item \ph{ticker\_signals}
\end{itemize}

\medskip
Here is the portfolio state:
\begin{itemize}[label=-, leftmargin=1.2em, itemsep=2pt, topsep=2pt]
  \item \ph{portfolio}
\end{itemize}

\medskip
The position ratio range: \br{0, \ph{max\_position\_ratio}}, the minimum step is 0.05.
\begin{itemize}[label=-, leftmargin=1.2em, itemsep=2pt, topsep=2pt]
  \item If you observe more bullish signals, you can set a larger position ratio.
  \item If you observe more bearish signals, you can set a smaller position ratio.
\end{itemize}

\medskip
You must provide your control recommendation as a structured output with the following fields:
\begin{itemize}[label=-, leftmargin=1.2em, itemsep=2pt, topsep=2pt]
  \item optimal\_position\_ratio: The optimal ratio of the position value to the total portfolio value
  \item justification: A brief explanation of your recommendation
\end{itemize}

\medskip
Your response should be well-reasoned and consider all aspects of the analysis.
\end{agentbox}

\section{table}

\begin{table*}[t!]
  \caption{Comparison of major CS2 skin trading platforms and their fee structures. We summarize representative selling and buyer fees, along with key features that influence effective transaction costs and liquidity, which motivate our choice of a 2\% friction level in the Transaction Friction Agent.}
  \centering
  \label{tradingruletb}
  \resizebox{.9\textwidth}{!}{
  \begin{tabular}{@{}lllp{0.4\linewidth}@{}}
  \toprule
  \textbf{Platform} & \textbf{Selling Fee} & \textbf{Buyer Fee} & \textbf{Key Features} \\ \midrule
  Steam Market(offical)  & 15\%          & N/A      & Official. Highest fees and funds are locked to the Steam Wallet (cannot be cashed out).           \\
  Buff        & 2.5\%          & 1\% Cash-out fee      & Highest liquidity globally. Uses P2P (Peer-to-Peer) trading.           \\
  YouPin        & 2\%-3\%       & 1\% Cash-out fee & Known for its ``Lease-to-Play'' system alongside standard P2P trading.        \\
  IGXE             & 2\%-3.5\%              & 1\% Cash-out fee    & Alternatives to Buff with occasional promotions and bot-delivery options.              \\
  CSFloat             & 2\%              & Varies (Low)    & Minimalist P2P site; excellent for high-tier items and specific "float" hunting.              \\
  Skinport             & 6\%-12\%              & 0\% for Buyers    & Extremely buyer-friendly; the listed price is the final price.              \\
  DMarket             & 5\%-7\%              & Varies    & Integrates trades for multiple games and blockchain assets.              \\
  SkinBaron             & 2\%-15\%              & Varies    & Very strict KYC/safety regulations. Competitive fees for items $\geq$ €1000.              \\ \bottomrule
  \end{tabular}
  }
\end{table*}

\section{Formal Definition of Evaluation Metrics}
\label{sec:metrics}

We evaluate each strategy using daily portfolio values and several standard financial metrics. Let $V_t$ denote the portfolio value at the end of trading day $t$, and $V_0$ be the initial capital. We define the simple daily return as
\begin{equation}
r_t = \frac{V_t - V_{t-1}}{V_{t-1}},
\end{equation}
and use it as the basis for all subsequent metrics.

\subsection{Cumulative Return (CR)}
The cumulative return over the full backtest horizon with $T$ trading days is defined as
\begin{equation}
\mathrm{CR} = \frac{V_T - V_0}{V_0} \times 100\%,
\end{equation}
which measures the total percentage change in portfolio value from the start to the end of the evaluation period.

\subsection{Sharpe Ratio (SR)}
We compute the Sharpe ratio using the excess daily return over a constant annual risk-free rate $r_f$ (set to 2\% in our experiments). First, we convert the annual risk-free rate into an effective daily rate under the standard 252 trading-day convention:
\begin{equation}
r_{f,d} = (1 + r_f)^{1/252} - 1.
\end{equation}
Let $\bar{r}$ be the mean of daily returns $r_t$, and $\sigma_r$ be their standard deviation. The annualized Sharpe ratio is then
\begin{equation}
\mathrm{SR} = \frac{\bar{r} - r_{f,d}}{\sigma_r} \times \sqrt{252},
\end{equation}
which measures the risk-adjusted return of the strategy.

\subsection{Annualized Volatility (AV)}
We use the standard deviation of daily returns to measure volatility. The annualized volatility (in percentage) is
\begin{equation}
\mathrm{AV} = \sigma_r \times \sqrt{252} \times 100\%,
\end{equation}
which reflects the magnitude of typical price fluctuations on an annualized basis.

\subsection{Maximum Drawdown (MDD)}
Maximum drawdown captures the worst peak-to-trough loss over the evaluation horizon. Let $M_t = \max_{1 \leq k \leq t} V_k$ be the running maximum of the portfolio value up to day $t$. The drawdown at day $t$ is
\begin{equation}
D_t = \frac{M_t - V_t}{M_t},
\end{equation}
and the maximum drawdown is
\begin{equation}
\mathrm{MDD} = \max_{1 \leq t \leq T} D_t \times 100\%,
\end{equation}
representing the largest percentage decline from a historical peak.

\subsection{Alpha ($\alpha$) and Beta ($\beta$)}
We estimate alpha and beta relative to the overall CS2 market index. Let $r_t^{(p)}$ denote the portfolio's daily return and $r_t^{(m)}$ the market index daily return. Using the same daily risk-free rate $r_{f,d}$ as above, we define excess returns
\begin{equation}
R_t^{(p)} = r_t^{(p)} - r_{f,d}, \quad R_t^{(m)} = r_t^{(m)} - r_{f,d}.
\end{equation}
We adopt the standard CAPM formulation
\begin{equation}
R_t^{(p)} = \alpha + \beta R_t^{(m)} + \varepsilon_t,
\end{equation}
and estimate $\alpha$ and $\beta$ via ordinary least squares over the backtest window. Equivalently,
\begin{equation}
\beta = \frac{\mathrm{Cov}(R^{(p)}, R^{(m)})}{\mathrm{Var}(R^{(m)})},
\end{equation}
\begin{equation}
\alpha = \overline{R^{(p)}} - \beta \; \overline{R^{(m)}},
\end{equation}
where $\overline{R^{(p)}}$ and $\overline{R^{(m)}}$ are the sample means of the excess returns. For presentation in Table~\ref{backboneres}, we report $\alpha$ in percentage terms. A positive $\alpha$ indicates excess return beyond what is explained by exposure to the overall market, while $\beta$ measures sensitivity to market movements (market risk).

\section{Construction of the CS2 Market Index}
\label{sec:csqaq_index}

We follow the CSQAQ index provider to construct a dimensionless overall CS2 market index that tracks the average price movement of a broad set of tradable items.

\subsection{Constituent Selection and Baseline}
Let $\mathcal{I}$ denote the set of items included in the index. The universe is constructed by selecting all items whose on-sale quantity on the platform exceeds a given liquidity threshold (greater than 50 listings at the time of construction), which yields $|\mathcal{I}| = N = 11{,}269$ items in our snapshot.

The index is anchored to a baseline date $t_0$ corresponding to 1 January 2021. On this date, the index level is fixed to
\begin{equation}
    I(t_0) = 1000.
\end{equation}
For each item $i \in \mathcal{I}$, let $P_i(t)$ denote its market price at date $t$. If an item was already listed on the baseline date, we take
\begin{equation}
    P_i(t_0)
\end{equation}
as its initial reference price. If an item was not yet listed on $t_0$, we use its first observed transaction or listing price as the initial reference price $P_i(t_0)$.

\subsection{Index Level at Time $t$}
At any later date $t$, the gross return of item $i$ relative to its initial reference price is
\begin{equation}
    R_i(t) = \frac{P_i(t)}{P_i(t_0)}.
\end{equation}
The index level is then defined as the equal-weighted average of these gross returns, scaled by the baseline value 1000:
\begin{equation}
    I(t) = 1000 \times \frac{1}{N} \sum_{i \in \mathcal{I}} R_i(t)
          = 1000 \times \frac{1}{N} \sum_{i \in \mathcal{I}} \frac{P_i(t)}{P_i(t_0)}.
\end{equation}
This construction removes the effect of different absolute price levels and puts all items on the same dimensionless scale before averaging.

\subsection{Rationale}
If we simply summed raw prices $\sum_i P_i(t)$ to form the index, very expensive items would dominate the index movement, and cheap items would have negligible influence. By normalizing each price by its initial reference price and then averaging, the index measures the average relative change across items rather than total market capitalization. As a result, the CS2 market index is not equal to the total market value or the sum of all item floor prices; instead, it is a dimensionless indicator that summarizes typical percentage changes in item prices over time.

\section{Additional Experimental Analysis}

\subsection{Performance Against Overall Market (Extended)}
Our system demonstrates high resilience in a declining market. During the test period, the overall CS2 market index fell by 15.62\%. Under the same conditions, our multi-agent framework significantly outperformed this benchmark across almost all backbone models. As reported in Table~\ref{backboneres}, Claude-sonnet-4 achieved the highest cumulative return of 7.58\%, while even the least successful model, Grok-4.1-fast-reasoning, experienced a loss of only 6.48\%. This confirms that the system is much more stable than the general market.

The market volatility was largely driven by specific official events. As illustrated in Figure~\ref{fig:assetvalue}, the CS2 ``Trade Up'' update on October 25 caused a massive market crash that wiped out a large portion of the skin market value. While the overall market struggled to recover, our multi-agent configurations showed a much faster and stronger recovery trend. This performance highlights that multi-agent systems are effective at navigating sudden, event-driven shocks in this specific market.

Unlike traditional securities markets, the CS2 market has a relatively small volume and shallow depth. Quantitative trading is not common here because of scale limits and platform-specific rules. In this environment, most trades are driven by individual emotions and sudden event news. Our system has a strong ability to collect information and perform reasoning, allowing it to adapt to the unique rules of the virtual asset market and obtain high ``alpha returns'', with the top-performing model achieving an alpha of 66.86\%.

\subsection{Agent-Level Findings (Extended)}
The CS2 market is a niche category where prices often do not reflect ``actual value'' due to high volatility and community bias. By testing different combinations of agents, we discovered several important market traits. Figure~\ref{fig:ablationvalue} and Table~\ref{ablationtb} show how each agent affects the outcome.

\etitle{Liquidity and the Realization Gap}:
By adding the Liquidity Agent, we observe a significant performance increase from 8923.05 to 10038.36 in final asset value. This agent improves the ``realization ability'' of the portfolio by preventing the system from buying items that are easy to list but hard to sell. The result highlights a specific market characteristic: many CS2 items suffer from a ``liquidity trap'' where listed prices are high but actual buyer demand is low.

\etitle{The Herd Effect in Community Sentiment}:
When we enable only the Sentiment Agent, its performance is poor. However, when we enable the Reversed Sentiment Agent, the final return jumps to 10593.76. Reversing community signals actually leads to better results than following them. This observation reveals that the CS2 market is heavily driven by a ``herd effect'', where widespread optimism often leads to ``market bubbles''. By identifying and countering this irrational hype, the system protects the portfolio from buying at artificial peaks.

\etitle{Official News as a Trading Inhibitor}:
After adding the Event Agent to the system, we observe a performance decrease from 10593.76 to 10383.21. Our investigation shows that official Steam updates often act as an ``inhibitor'' of profit. Official news is a form of ``common knowledge'' that is quickly priced in by all participants, and therefore contains little unique profit opportunity for a trading system. This demonstrates that the CS2 market is highly efficient with respect to official news, making these signals less useful for high-return strategies.

\etitle{The Impact of Transaction Friction}:
The data shows a large difference between ``ideal'' returns and ``realistic'' returns. When we do not consider trading costs, the profit appears much higher (12196.48). However, once we include the 2\% transaction fee, the return is about 14\% lower at 10593.76. This confirms that transaction fees cannot be ignored in CS2 trading. These costs do more than simply erode profit: they directly influence the decision-making process. If a model ignores fees, it may suggest frequent, small trades that appear profitable on paper. In contrast, the Transaction Friction Agent recognizes that a 2\% fee can turn a small gain into a net loss, forcing the system to be more selective and often to choose ``Hold'' instead of inefficient ``Sell'' actions.

\subsection{Backbone Behavior (Extended)}
We also examine how different LLM backbones behave under the same CSTrader framework. The models fall into four broad categories based on their risk preferences and analytical focus.

\etitle{Smart Models:}
The first category includes models like \textbf{Claude-sonnet-4}, which are effective at identifying specific market opportunities after a crisis. During the early stage of the experiment, Claude maintains a cautious stance with high cash reserves, resulting in performance that stays near the break-even line. After the market crash on October 25, the model accurately identifies a recovery opportunity. By increasing its position size at the optimal time, Claude achieves the highest CR of 7.58\% and the highest $\alpha$ of 66.86\%, showing a strong ability to find ``alpha returns'' during periods of high uncertainty.

\etitle{Stable and Risk-Averse Models:}
The second category is represented by \textbf{Qwen-Max}, which follows a conservative and disciplined style. This model prioritizes risk management and maintains strict rules for entering and exiting positions. Its asset value curve is the most stable among all profitable models, with the lowest AV of 38.86\% and the lowest $\beta$ of 0.20. Although its final CR of 3.83\% is not the highest, its high SR of 0.42 demonstrates that it offers the best risk-adjusted stability in a volatile market.

\etitle{Aggressive Models:}
Models like \textbf{DeepSeek-V3.2} exhibit an aggressive trading style with high capital utilization. Early in the experiment, DeepSeek maintains near-full positions, which leads to high performance during the flourishing phase. However, this aggressive style also results in the highest AV of 106.10\% and the highest $\beta$ of 0.65. During the crash on October 25, DeepSeek suffers the largest losses and ranks near the bottom for several days. Although it partially recovers afterward, its final CR remains negative at -2.89\%, indicating that aggressive backbones struggle with risk control during sudden shocks.

\etitle{Weak Analysis and Selection Models:}
The final category includes models like \textbf{Gemini-3-flash-preview}, which show weak analysis of market potential and item dynamics. Although Gemini increases its position size after the market dip, its final CR of -5.65\% is underwhelming. The model appears to struggle in selecting the correct assets within the CS2 market; even when it approximately times the market, the chosen items do not gain enough value to offset previous losses. This indicates that effective trading in this niche market requires not only timing but also a strong ability to identify which specific items have the highest growth potential.

\end{document}